%% file: main.tex
\documentclass[10pt,twocolumn,letterpaper]{article}
\pdfoutput=1

\usepackage{float}
\usepackage[pagenumbers]{iccv} 


\usepackage[pagebackref,breaklinks,colorlinks]{hyperref}
\usepackage[ruled,linesnumbered]{algorithm2e}
\usepackage{amssymb}
\usepackage{bbding}
\def\textbi#1{\textbf{\textit{#1}}}

\title{Towards Seamless Borders: A Method for Mitigating Inconsistencies\\ in Image Inpainting and Outpainting}

\author{
	\centerline{Xingzhong Hou$^{1,2*}$, Jie Wu$^{1*}$, Boxiao Liu$^{1}$, Yi Zhang$^{1}$, Guanglu Song$^{1}$, Yunpeng Liu$^{1,2}$, Yu Liu$^{1}$, Haihang You$^{2}$} \\
	\centerline{$^1$ Sensetime Research \quad $^2$ Institute of Computing Technology, Chinese Academy of Sciences} \\
	\centerline{\tt\small{\{houxingzhong, liuyunpeng22b, youhaihang\}@ict.ac.cn}} \\
	\centerline{\tt\small{\{wujie.cs, liuyuisanai\}@gmail.com}} \\
	\centerline{\tt\small{\{liuboxiao, zhangyi17, songguanglu\}@sensetime.com}}
}

\begin{document}
\maketitle
\def\thefootnote{*}\footnotetext{Equal Contribution}

\begin{abstract}
	
	Image inpainting is the task of reconstructing missing or damaged parts of an image in a way that seamlessly blends with the surrounding content. With the advent of advanced generative models, especially diffusion models and generative adversarial networks, inpainting has achieved remarkable improvements in visual quality and coherence. However, achieving seamless continuity remains a significant challenge. In this work, we propose two novel methods to address discrepancy issues in diffusion-based inpainting models. First, we introduce a modified Variational Autoencoder that corrects color imbalances, ensuring that the final inpainted results are free of color mismatches. Second, we propose a two-step training strategy that improves the blending of generated and existing image content during the diffusion process. Through extensive experiments, we demonstrate that our methods effectively reduce discontinuity and produce high-quality inpainting results that are coherent and visually appealing.
	
\end{abstract}

\section{Introduction}

Image inpainting~\cite{xu2023review,suvorov2022resolution}, which involves filling in missing or obscured parts of an image, has long been a challenging problem in computer vision. Recent advancements in deep learning have led to the development of generative models capable of producing high-quality inpainted images~\cite{hui2020image, liu2020rethinking, ntavelis2020aim, ho2020denoising,song2020denoising,Rombach_2022_CVPR, karras2020analyzing}, with content that blends seamlessly with the surrounding regions. Among these models, diffusion models~\cite{ho2020denoising,song2020denoising,Rombach_2022_CVPR} have gained prominence for image synthesis and inpainting, utilizing a denoising process to gradually convert noisy data into detailed, high-resolution images.

Despite the effectiveness of recent diffussion-based methods~\cite{avrahami2023blended, manukyan2023hd, ju2024brushnet, zhuang2023task} in image inpainting and outpainting,  a critical challenge remains largely overlooked, yet is crucial in real-world applications. In practical scenarios, when a user inpaints an image, it is essential that \textbf{the unmasked areas remain unaltered}.
However, existing approaches typically generate the entire output image, including the unmasked regions, which can inadvertently modify the unmasked areas.
This results in a loss of original information, as illustrated in \cref{fig:intro}, where elements like human faces or letters become distorted. A straightforward solution is to directly paste the original image onto the unmasked regions of the output. While this may preserve the original content, it often leads to noticeable mismatches at the boundaries between the inpainted and unmasked areas, both in terms of color and content. As show in \cref{fig:intro}, the background of inpainting method exhibits noticeable boundaries along the mask edges.

We analyze the standard approach for diffusion-based inpainting/outpainting and identify that the primary cause of boundary inconsistencies lies in the Variational Autoencoder (VAE)~\cite{kingma2013auto} and the training process of the diffusion model itself.
In the standard inpainting pipeline, the latent representations of masked and unmasked regions are blended at each denoising step to maintain coherence. Although this blending approach helps, it is insufficient to guarantee complete consistency. The VAE, trained on unmasked images for general image reconstruction, struggles to accurately capture the context of masked images. This misalignment arises because the VAE was not explicitly trained to handle partially masked images, and thus it often fails to generate coherent representations along the boundaries of masked regions. In the process of training diffusion models, most methods overlook the blending operation, which creates an inconsistency with the blending that occurs during inference.

To address these limitations, we propose two complementary solutions. First, we fine-tune the VAE to adapt to the inpainting task by training it to handle masked images directly. By modifying the VAE, we ensure that the reconstructed images remain color continuous across masked regions. Second, we introduce a two-step training paradigm for the diffusion model, where we simulate the blending operation during training. This approach enables the model to learn to handle masked and unmasked regions more coherently throughout the diffusion process, resulting in more reasonable structure. Our experiments demonstrate that these methods effectively mitigate discontinuities and improve the perceptual quality of inpainted/outpainted images. 

Our main contributions are as follows:

\begin{itemize}
	\item We identify and analyze a critical issue in diffusion-based inpainting/outpainting tasks, where discontinuities emerge at the boundaries between the generated and original image regions.
	\item We introduce a refined VAE model specifically tailored for inpainting/outpainting tasks, ensuring that the training objective of the diffusion model is continuous in the latent space and the color is smooth along the edge of mask in the image space.
	\item We propose a novel two-step training paradigm, which simulates the blending operation in the training stage to enhance the consistency between the generated content and the original image.
	\item Extensive qualitative and quantitative results demonstrate that our proposed method substantially reduces discontinuities and improves overall inpainting/outpainting quality when compared to existing image inpainting/outpainting methods.
\end{itemize}

\begin{figure*}[t]
	\centering
	\includegraphics[width=\linewidth]{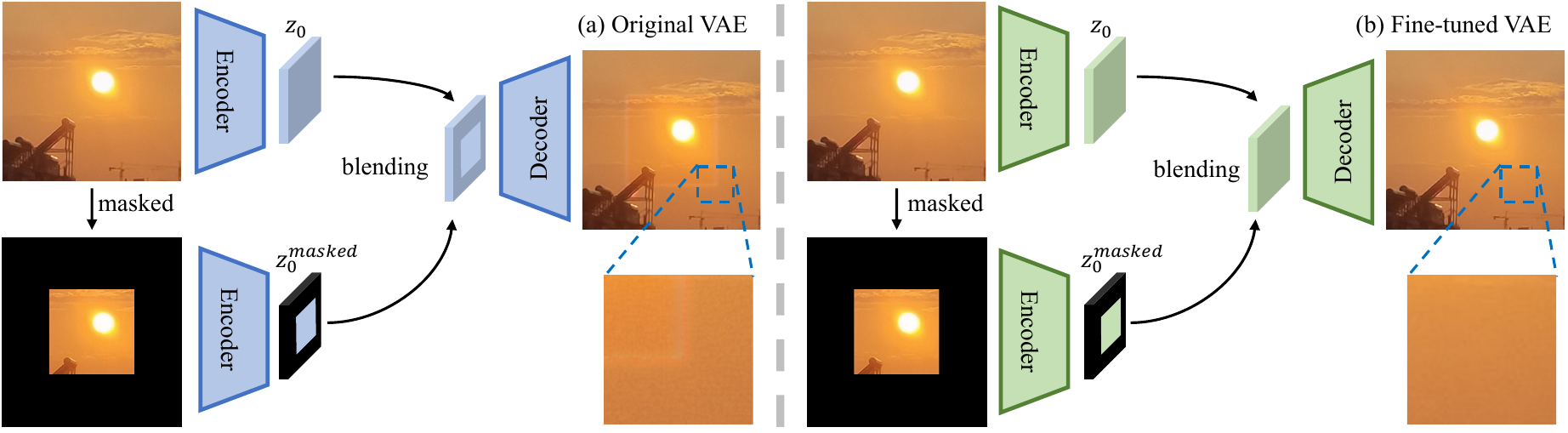}
	\caption{The reconstruction results of the VAE under the blending operation. It can be observed that the original VAE produces noticeable color discontinuity in the reconstructed blended result, whereas our fine-tuned VAE achieves a much smoother reconstruction, almost identical to the original image.}
	\label{fig:VAE}
\end{figure*}

\section{Related work}
The image inpainting task is to reasonably fill a user-specific area in an image, which involves erasing objects from selected areas or expanding the image beyond a specified region. Traditional methods are based on Variational Autoencode (VAE)~\cite{kingma2013auto, peng2021generating, zheng2019pluralistic, zhao2020uctgan} and generative adversarial networks (GANs)~\cite{goodfellow2014generative, sargsyan2023mi, xu2023image,zhao2021large, zheng2203cm}. These methods typically employ specific loss functions to generate realistic inpainting structures and often require carefully designed network architectures to enhance content-consistent quality. However, the generalization of these methods often only targets specific fields and has poor generalization. 
In recent years, diffusion models~\cite{ho2020denoising,song2020denoising,Rombach_2022_CVPR, peebles2023scalable, yang2023diffusion} have achieved significant progress in generative modeling tasks and have been successfully applied to image inpainting~\cite{saharia2022palette, lugmayr2022repaint}. Compared to traditional methods, diffusion models generate high-quality images with natural textures and structures in masked regions by employing a gradual denoising process.

Blend Latent Diffusion~\cite{avrahami2023blended} is a representative diffusion-based inpainting method, which enables inpaint specified masked regions by replacing non-edited areas with the original image at each denoising step in the latent space, without altering the pre-trained diffusion model. This method has become a widely adopted paradigm in current diffusion based inpainting approaches. SD-Inpainting is a fine-tuned version of the original SD model. It modifies the input channels of the diffusion model by channel-wise concatenating the masked image condition and the mask in the latent space. SmartBrush~\cite{xie2023smartbrush} improve the shape-guided inpainting by introducing multiple masks of the same object during training. HD-Painter~\cite{manukyan2023hd} enhances the text alignment in the painting region by introducing Prompt-Aware Introverted Attention. BrushNet~\cite{ju2024brushnet} designs a dual-branch network that separates the masked image features and noisy latent vectors into different branches. The features from the branch networks are injected into the frozen pre-trained main network layer by layer, which increases the consistency of the generation and enhances the inpainting effect. PowerPaint~\cite{zhuang2023task} integrates multiple inpainting tasks into one model by introducing learnable task prompts and targeted fine-tuning strategies to achieve multiple image editing tasks.

Although these methods have shown promising results in structural coherence and diversity for content filling, they still display noticeable discrepancies and boundary artifacts when blending non-generated image regions with generated content in the image space. This limitation restricts their practical applications, where it is often essential for non-masked regions to remain entirely unchanged.
\section{Method}

\subsection{Preliminaries}

Denoising Diffusion Probabilistic Models (DDPM)~\cite{ho2020denoising} aim to transform pure noise $z_T$ into a coherent output image $z_0$, guided by the given conditioning prompt. In the forward process, noise is added to a clean image, which can be represented as:

\begin{equation}
	z_t = \sqrt{\bar{\alpha_t}}z_0 + \sqrt{1-\bar{\alpha_t}}\epsilon
\end{equation}

where $z_t$ denotes the noisy sample at diffusion step $t$, and $\bar{\alpha_t}$ are hyper-parameters governing the noise schedule over $t \in [1, T]$. 

During training, the network $\theta$ is optimized to predict the noise $\epsilon$ given the noisy latent variable $z_t$:

\begin{equation}
	\mathcal{L}_{\theta} = \mathbb{E}_{z_0, \epsilon \thicksim N(0,I), t \thicksim U(1,T)} \| \epsilon - \epsilon_{\theta}(z_t, t, \mathcal{C}) \|_2^2,
\end{equation}

where $\mathcal{C}$ is the conditioning input to $\theta$. To enhance performance in low-SNR steps, v-prediction~\cite{salimans2022progressive} is employed, where the model targets the variable $\textbf{v}$, formulated as:

\begin{equation}
	\textbf{v} = \sqrt{\bar{\alpha_t}}\epsilon - \sqrt{1-\bar{\alpha_t}}z_0.
\end{equation}

The original image $z_0$ can then be approximated by:

\begin{equation}
	\hat{z_0} = \sqrt{\bar{\alpha_t}}z_t - \sqrt{1-\bar{\alpha_t}}\textbf{v}_{\theta}(z_t)
\end{equation}

Stable Diffusion (SD)~\cite{Rombach_2022_CVPR} further incorporates a Variational Auto-Encoder (VAE)~\cite{kingma2013auto} to map the input image into a lower-dimensional latent space. In this setting, the latent representation $z_0$ is obtained as $E(x_0)$, where $x_0$ is the input image, and the reconstructed output image is derived as $D(z_0)$. Our work builds upon the Stable Diffusion model architecture.

\begin{figure*}[t]
	\centering
	\includegraphics[width=0.85\linewidth]{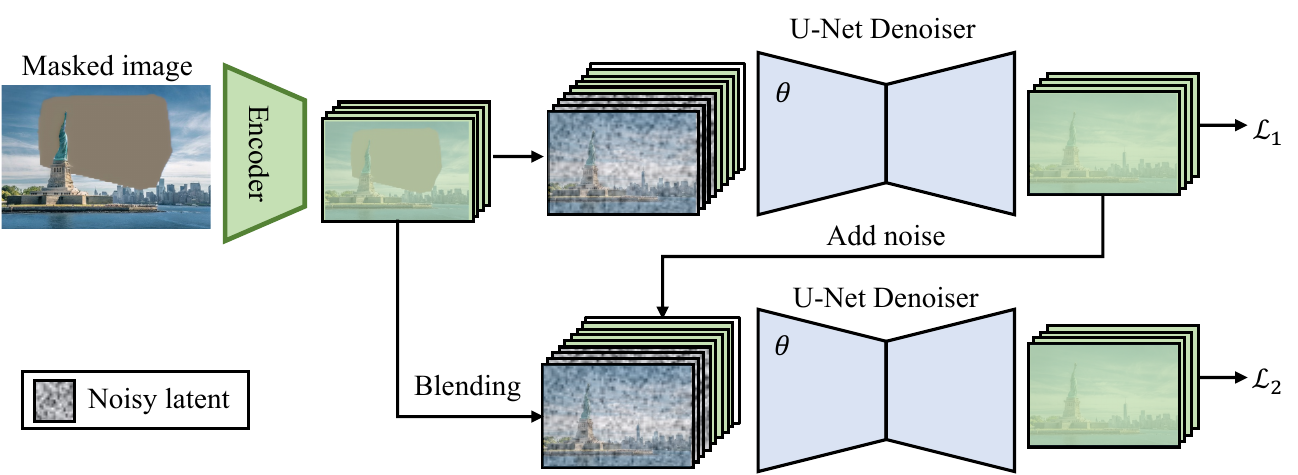}
	\caption{The illustration of our proposed two-step training paradigm. In the first step, $\mathcal{L}_{1}$ represents the standard diffusion model loss, where the model predicts
		$z_0$ normally. After obtaining $z_0$, we simulate the blending operation and calculate the latent code for the next timestep. This updated latent code then serves as the input for the next step, where the model predicts the noise to calculate $\mathcal{L}_{2}$, effectively simulating the blending operation throughout the training process. The $\theta$ in the figure represents the same denoiser during training.}
	\label{fig:pipeline}
\end{figure*}

\subsection{Discontinuity in inpainting/outpainting}
\label{Color discontinuity}

For inpainting/outpainting tasks, Blended Latent Diffusion (BLD)~\cite{avrahami2023blended} is one of the most widely used methods, often serving as the default inference approach in diffusion-based inpainting/outpainting models. Given a binary mask $m$ and an original image $x_0$, the masked image is defined as $x_0^{masked} = x_0 * m$, where the areas to be inpainted/outpainted in $x_0$ have a mask value of 0, and the unchanged regions have a mask value of 1. 
BLD begins by extracting the latent representation $z_0^{masked}$ from $x_0^{masked}$ using VAE. Then, the binary mask $m$ is resized to match the shape of $z_0^{masked}$, creating $m^{resized}$ to ensure compatibility in subsequent operations. To maintain consistency in the unchanged regions throughout the diffusion process, BLD applies noise to $z_0^{masked}$ up to the desired noise level at time step $t$, yielding $z_t^{masked}$. At each diffusion step, the latents $z_t$ (the current state of inpainting/outpainting) and $z_t^{masked}$ (the noisy masked latent) are blended using the resized mask as follows:

\begin{equation}
	\label{eq:1}
	z_t = z_t * (1 - m^{resized}) + z_t^{masked} * m^{resized}.
\end{equation}

The blending operation ensures that the inpainting/outpainting remains true to the original image in the preserved regions, while allowing for new content generation in the masked areas.

At each denoising step, the entire latent representation is modified, but subsequent blending ensures that the regions inside $m^{resized}$ remain unchanged. However, this blending approach does not guarantee coherence, as $z_t^{masked}$ is derived from the VAE with $x_0^{masked}$, while the diffusion model is trained to target the latent representation of the whole image. In particular, to keep the regions inside $m$ consistent with the original, blending is applied not only during the denoising steps but also to the final image generated by the VAE’s decoder. As illustrated in \cref{fig:VAE}, the blended output $\hat{x} = D(z_0 * (1 - m^{resized}) + z_0^{masked} * m^{resized})$ shows noticeable color discontinuities along the edges of mask, indicating a mismatch between the inpainting target and the actual diffusion model objective. Additionally, the absence of blending during training impacts content consistency in generation, as it introduces a mismatch with the blending operation applied during inference.

To address the discontinuity, we propose two methods: VAE fine-tuning and a two-step training paradigm, which are detailed in \cref{VAE fine-tuning} and \cref{Two steps training paradigm}.

\begin{algorithm}[t]
	\caption{Two-step training paradigm}
	\label{alg:1}
	\KwIn{Network $\theta$, Noises $\epsilon_1$ $\epsilon_2$, Sample $\textbi{z}_0$, Mask $m$, Timestep $t$}
	\KwOut{Optimized network $\theta$}
	\BlankLine
	
	$z_t = \sqrt{\bar{\alpha_t}}z_0 + \sqrt{1-\bar{\alpha_t}}\epsilon_1$;
	
	$v_1 = \sqrt{\bar{\alpha_t}}\epsilon_1 - \sqrt{1-\bar{\alpha_t}}z_0$;
	
	$\mathcal{L}_{1} = \| \textbf{v}_1 - \textbf{v}_{\theta}(z_t, t, \mathcal{C}) \|_2^2$;
	
	$\hat{z}_0 = \sqrt{\bar{\alpha_t}}z_t - \sqrt{1-\bar{\alpha_t}}\textbf{v}_{\theta}(z_t)$;
	
	Blending operation $\hat{z_0} = \hat{z_0} * (1 - m^{resized}) + z_0^{masked} * m^{resized}$;
	
	$\hat{z_{t-1}} = \sqrt{\bar{\alpha_{t-1}}}\hat{z_0} + \sqrt{1-\bar{\alpha_{t-1}}}\epsilon_2$;
	
	Calculate the true noise $\epsilon_2^* = \frac{\hat{z_{t-1}}-\sqrt{\bar{\alpha_{t-1}}}z_0}{\sqrt{1-\bar{\alpha_{t-1}}}}$;
	
	$\textbf{v}_2 = \sqrt{\bar{\alpha_{t-1}}}\epsilon_2^* - \sqrt{1-\bar{\alpha_{t-1}}}z_0$;
	
	$\mathcal{L}_{2} = \| \textbf{v}_2 - \textbf{v}_{\theta}(z_{t-1}, t, \mathcal{C}) \|_2^2$;
	
	$\theta \leftarrow \theta - \eta \nabla_{\theta}(\mathcal{L}_{1} + \lambda\mathcal{L}_{2})$;
	
	\Return $\theta$
	
\end{algorithm}

\begin{figure*}[t]
	\centering
	\includegraphics[width=\linewidth]{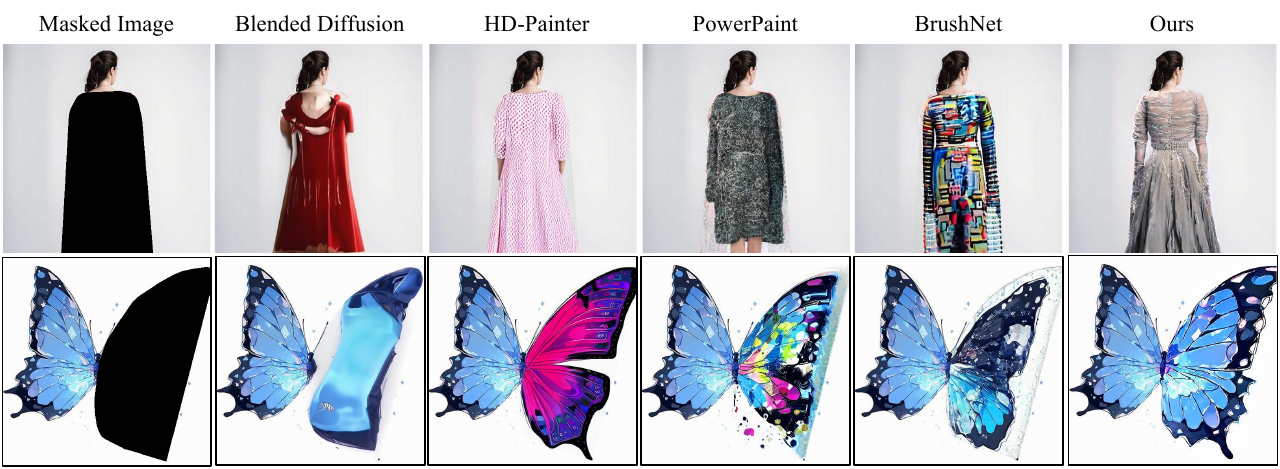}
	\caption{Comparison of color continuity between our method and other approaches for inpainting.}
	\label{fig:comparison_color_inpainting}
\end{figure*}

\subsection{Refined VAE for inpainting/outpainting}
\label{VAE fine-tuning}

As discussed in \cref{Color discontinuity}, blending operations in latent space can lead to discontinuities, as the VAE, trained to reconstruct complete images, is not inherently suited for handling images with masked regions. As shown in \cref{fig:VAE}, original VAE produces color discontinuous results with blending opreation. In the VAE architecture, consisting of an encoder $E$ and a decoder $D$, standard image reconstruction is formulated as $\hat{x} = D(E(x))$. However, in inpainting/outpainting tasks, due to blending at each timestep, the latent $z_t$ is updated as shown in \cref{eq:1}, and the reconstruction target of VAE is shifted to:

\begin{equation}
	\label{eq:2}
	\hat{x} = D(E(x) * (1 - m^{resized}) + E(x * m) * m^{resized}).
\end{equation}

To achieve consistent image reconstruction, the pre-trained VAE is fine-tuned based on the modified reconstruction objective in \cref{eq:2}. With the refined VAE, the target diffusion model, $z_0$, is continuous in the latent space of VAE, and the inpainting/outpainting results will be generated seamlessly, free from color discontinuity.

\subsection{Two-step training paradigm}
\label{Two steps training paradigm}

The refinement of the VAE resolves color discontinuity in the latent space, but the challenge of content consistency during the diffusion generation process remains. To address this, we propose a two-step training paradigm that simulates the blending operation directly during training.

In the conventional v-prediction diffusion training process, the objective is to predict the velocity $\textbf{v}$ from the latent variables $z_t$:

\begin{equation}
	\mathcal{L}_{\theta} = \| \textbf{v} - \textbf{v}_{\theta}(z_t, t, \mathcal{C}) \|_2^2,
\end{equation}

After the model predicts the noise, we further infer $\hat{z_0}$ and $\hat{z_{t-1}}$, applying a blending operation on $\hat{z_{t-1}}$ to replicate the inference procedure. As detailed in \cref{alg:1}, we compute $\hat{z_0}$ using the predicted $\textbf{v}_{\theta}$, followed by a blending operation to update $\hat{z_0}$. It’s important to note that, while the blending operation in inference is applied to $z_t$, during training, it’s performed on $\hat{z_0}$. Since $\hat{z_{t-1}}$ is calculated by adding noise to $\hat{z_0}$, blending $\hat{z_0}$ is effectively equivalent to blending $\hat{z_{t-1}}$.

Given that $\hat{z_0}$ is not a real sample, we compute the true noise $\epsilon_2^*$ to adjust the inpainting/outpainting accordingly. With $\epsilon_2^*$, we derive the target $\textbf{v}_2$ for the second step. Finally, the model parameters $\theta$ are updated based on the combined losses $\mathcal{L}_{1} + \lambda\mathcal{L}_{2}$, where $\lambda$ serves as a weight factor to balance the two loss terms. The two-step training paradigm effectively enhances content consistency by integrating the blending simulation during training, resulting in smoother and more consistent inpainting/outpainting outputs.

\section{Experiments}

In this section, we evaluate the proposed method qualitatively and quantitatively, for both inpaiting and outpainting.

\begin{figure*}[t]
	\centering
	\includegraphics[width=\linewidth]{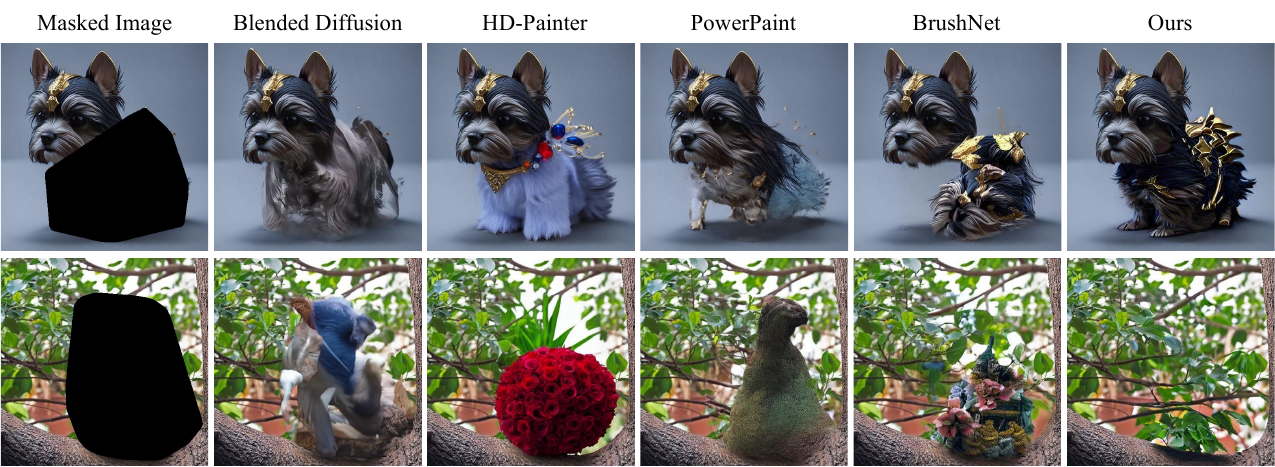}
	\caption{Comparison of content consistency between our method and other approaches for inpainting.}
	\label{fig:comparison_content_inpainting}
\end{figure*}

\paragraph{Implementation details.} In our experiments, unless otherwise specified, all tasks are based on Stable Diffusion 1.5 (SD-1.5). The modified VAE is trained on 256×256 resolution images using the default training settings, consistent with those in the SD-1.5 VAE training phase. Apart from splitting the original VAE input into two parts, the VAE model structure remains unchanged. Specifically, we merged the two input parts in the latent space and reconstructed the result after merging. For denoiser, we set the number of the sampling steps to be 50 via using the stride of 20 over 1000 diffusion steps with a guidance scale of 3. The loss weight $\lambda$ is set to 0.5 by default. For the inpainting task, we utilize the saliency detection model U2Net~\cite{qin2020u2} to segment out the background, then randomly sample masks within the background areas as input during training. For outpainting, we apply a mask around the edges of the training images, retaining only the central portion. In both tasks, the training objective of denoiser is to reconstruct the original training images. 

\paragraph{Dataset.} To train the inpainting/outpainting model, we filter the publicly available Laion-Aesthetic~\cite{schuhmann2022laion} dataset, selecting images with an aesthetic score of 6.0 or higher and a resolution of at least 512x512 as our training set. Both the VAE and denoiser are fine-tuned based on this dataset. 

\paragraph{Evaluation benchmarks.} We employ BrushBench introduced in BrushNet~\cite{ju2024brushnet} as our evaluation dataset as it comprises various types of images. Since the inpainting masks provided by BrushNet are derived from segmentation results and thus contain strong prior information, we make some modifications to these masks. For each inpainting mask, we first identify its convex hull and then slightly expand this convex hull to create a new mask. Through this process, the new mask no longer retains the distinct shape of the segmented object, and it is used for inpainting tasks. For those inpainting tasks guided by prompts, we use `scenery', which is the best one shown in PowerPaint, as the default prompt.

\begin{table}[t]
	\centering
	\begin{tabular}{l c c c}
		\hline
		Metrics & \multicolumn{2}{c}{Image Quality} & Consistency \\
		\hline
		Models & FID($\downarrow$) & AS($\uparrow$) & CD($\downarrow$) \\
		\hline
		BLD~\cite{avrahami2023blended} & 18.466 & 5.367 & 15.735 \\
		HD-Painter~\cite{manukyan2023hd} & 15.604 & 5.985 & 19.029 \\
		PowerPaint~\cite{zhuang2023task} & 15.848 & 5.550 & 12.754 \\
		BrushNet~\cite{ju2024brushnet} & 17.824 & 5.859 & 21.036 \\
		Ours  & \textbf{15.434} & \textbf{5.991} & \textbf{8.538} \\
		\hline
	\end{tabular}
	\caption{Quantitive evaluation of different inpainting methods based on BrushBench. Evaluation Metrics include image quality and blending consistency.}
	\label{tab:comparison}
\end{table}

\paragraph{Evaluation metrics.} Our evaluation metrics focus on two main aspects to assess the generated results: image quality and blending consistency. We utilize two metrics to measure the quality of the generated images: Fréchet Inception Distance (FID)~\cite{heusel2017gans}, and Aesthetic Score (AS)~\cite{schuhmann2022laion}, each applied according to their default configurations. Specifically, FID is used for image visual quality. Aesthetic Score is a linear model trained on image quality rating pairs of real images. We measure color continuity by calculating the L2 distance of RGB values between pixels along the mask boundary and their neighboring pixels, called Color Distance (CD).

\begin{table}[t]
	\centering
	\begin{tabular}{l c c c}
		\hline
		Metrics & \multicolumn{2}{c}{Image Quality} & Consistency \\
		\hline
		Models & FID($\downarrow$) & AS($\uparrow$) & CD($\downarrow$) \\
		\hline
		HD-Painter~\cite{manukyan2023hd} & 30.596 & 5.897 & 14.871 \\
		PowerPaint~\cite{zhuang2023task} & 22.768 & 5.606 & 12.389 \\
		BrushNet~\cite{ju2024brushnet} & 30.342 & 5.487 & 15.458 \\
		Ours  & \textbf{19.530} & \textbf{6.131} & \textbf{10.249} \\
		\hline
	\end{tabular}
	\caption{Quantitive evaluation of different methods for outpainting tasks based on BrushBench.}
	\label{tab:comparison_outpainting}
\end{table}

\begin{figure*}[t]
	\centering
	\includegraphics[width=0.8\linewidth]{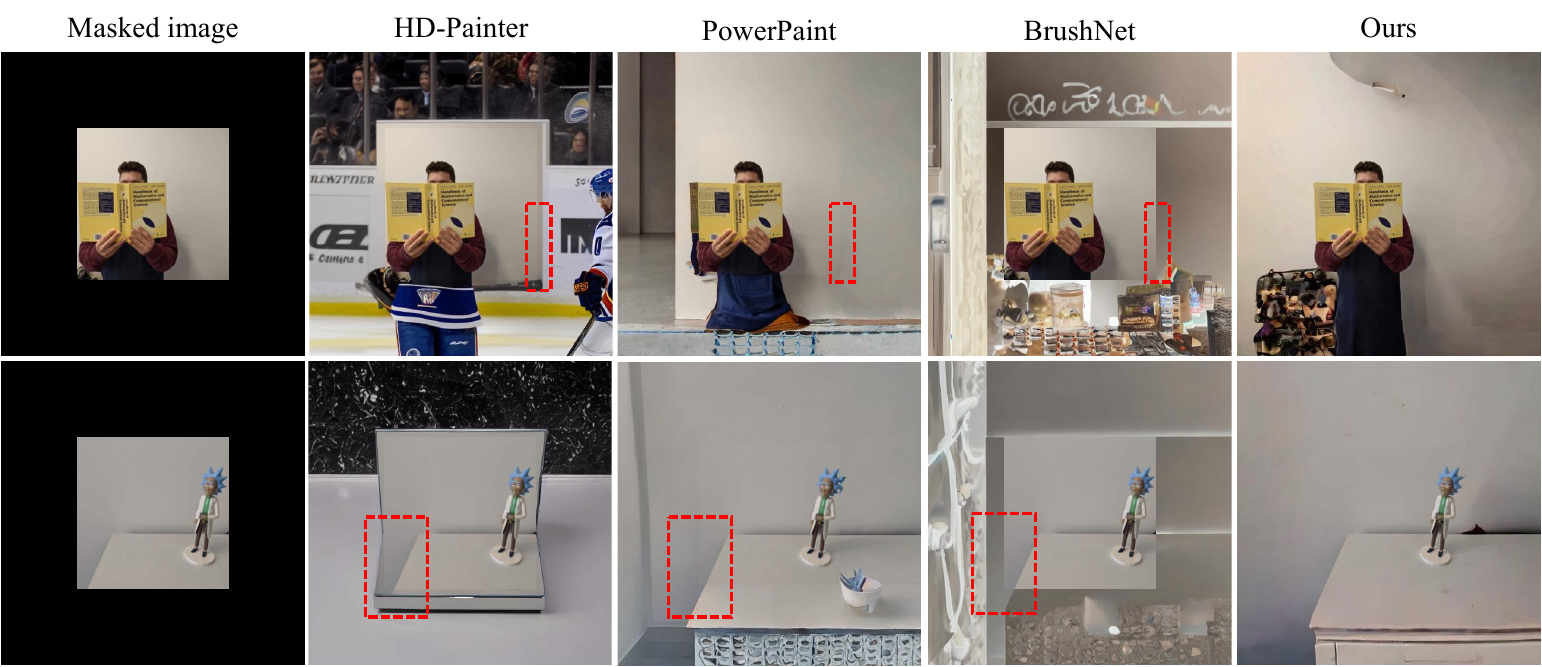}
	\caption{Comparison of color continuity between our method and other approaches for outpainting.}
	\label{fig:comparison_color_outpainting}
\end{figure*}

\begin{figure*}[t]
	\centering
	\includegraphics[width=0.8\linewidth]{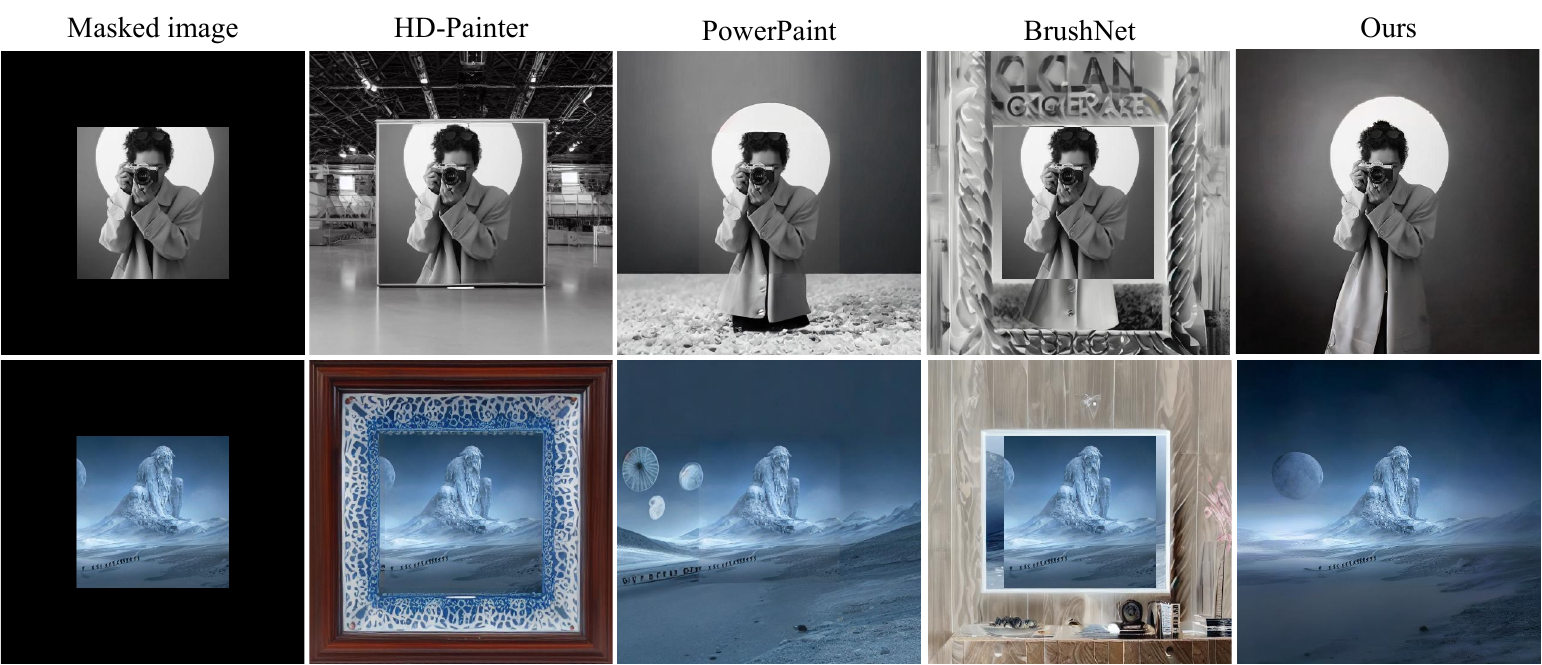}
	\caption{Comparison of content consistency between our method and other approaches for outpainting.}
	\label{fig:comparison_content_outpainting}
\end{figure*}

\subsection{Qualitative evaluation}

We conduct extensive experiments for the qualitative comparison between our method and SOTA approaches. We keep the recommended hyper-parameter for each inpainting method in all images for fair comparison.

We present the visual results of different methods on the inpainting task, from two perspectives, color continuity and content consistency, as shown in \cref{fig:comparison_color_inpainting} and \cref{fig:comparison_content_inpainting} respectively. In \cref{fig:comparison_color_inpainting}, existing inpainting methods always exhibit clear color boundaries, while our method demonstrates excellent color continuity. In inpainting tasks, reasonable structure is crucial to create a cohesive, visually complete image. Our method generates images with more coherent and realistic structures than other methods, which is presented in ~\cref{fig:comparison_content_inpainting}.

We also visualize the results of different methods on the outpainting task in ~\cref{fig:comparison_color_outpainting} and ~\cref{fig:comparison_content_outpainting}. As seen in ~\cref{fig:comparison_color_outpainting}, in the transition area between the non-masked and generated regions, the compared methods show noticeable color discontinuity, whereas our results do not suffer from this issue. As shown in ~\cref{fig:comparison_content_outpainting}, in terms of content structure generation, BrushNet and HD-painter tend to produce frames around the image, likely due to the lack of specific fine-tuning for the outpainting task during training, which involves in PowerPaint. Compared to PowerPaint, our method demonstrates advantages in structure coherence, such as the generation of human body structures in the first row and the moon in the second row. Consequently, for both inpainting and outpainting tasks, our approach demonstrates clear advantages in both structure and color coherence.

\subsection{Quantitative evaluation} 

We conduct a comprehensive quantitative evaluation by comparing our method to SOTA diffusion-based inpainting approaches. The results, presented in \cref{tab:comparison}, clearly demonstrate that our approach significantly outperforms other inpainting methods across multiple evaluation metrics. Specifically, our method achieves superior results in both image quality and consistency, effectively reducing visible artifacts at the boundary of the inpainted regions. Furthermore, we extend our evaluation to outpainting tasks, with quantitative comparisons shown in \cref{tab:comparison_outpainting}. The results validate our method’s ability to handle larger context expansions beyond the masked regions, demonstrating robustness and flexibility across both inpainting and outpainting applications.

\begin{figure}[t]
	\centering
	\includegraphics[width=\linewidth]{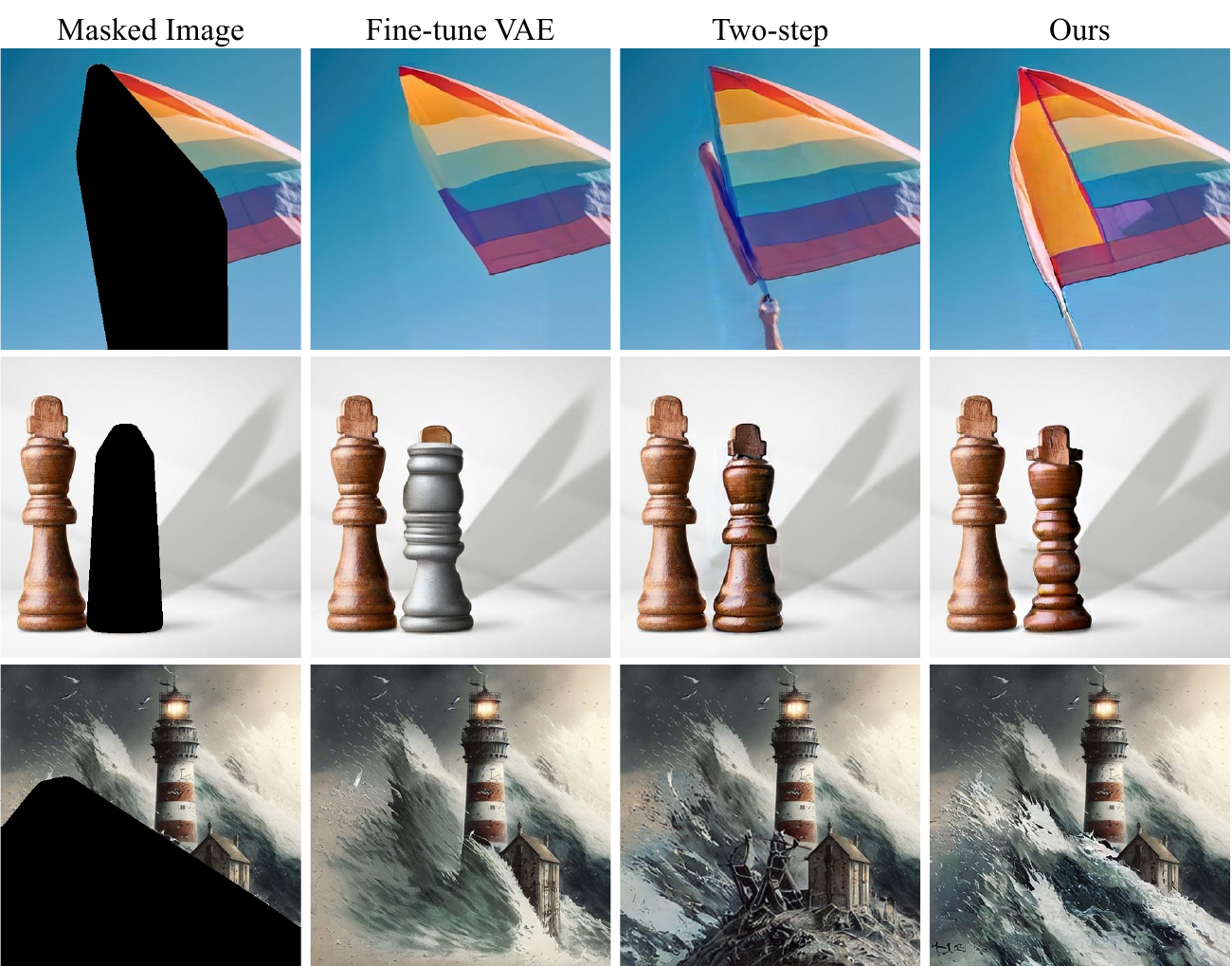}
	\caption{Ablation study for inpainting.}
	\label{fig:ablation_inpainting}
\end{figure}

\subsection{User study}

To further evaluate the effectiveness of our method in producing visually coherent and seamless inpainting and outpainting results, we conducted a user study. The study involved 20 participants. We randomly sample test images with all methods and ask them to choose the most satisfying results per different target. Everyone evaluates 100 images, 50 images for inpainting and 50 images for outpainting. The results are shown in \cref{tab:user_study}. It shows that our results are chosen mostly frequent in both tasks.

\begin{table}[t]
	\centering
	\begin{tabular}{l c c}
		\hline
		& Inpainting & Outpainting \\
		\hline
		BLD~\cite{avrahami2023blended} & 2\% & - \\
		HD-Painter~\cite{manukyan2023hd} & 32\% & 4\% \\
		PowerPaint~\cite{zhuang2023task} & 10\% & 14\% \\
		BrushNet~\cite{ju2024brushnet} & 10\% & 0\% \\
		Ours  & \textbf{46\%} & \textbf{82\%} \\
		\hline
	\end{tabular}
	\caption{A user study to show the comparison between our method and other approaches.}
	\label{tab:user_study}
\end{table}

\subsection{Ablation study}

In our method, both the VAE refinement and the two-step training paradigm are designed to enhance image consistency. The fine-tuned VAE primarily addresses color continuity, ensuring that the generated region blends smoothly with the original image. On the other hand, the two-step training paradigm ensures content consistency, maintaining structural coherence between the generated and original contents. To validate the effectiveness of these components, we conduct a series of ablation studies, demonstrating the individual contributions to the overall performance of our method.

\begin{figure}[t]
	\centering
	\includegraphics[width=\linewidth]{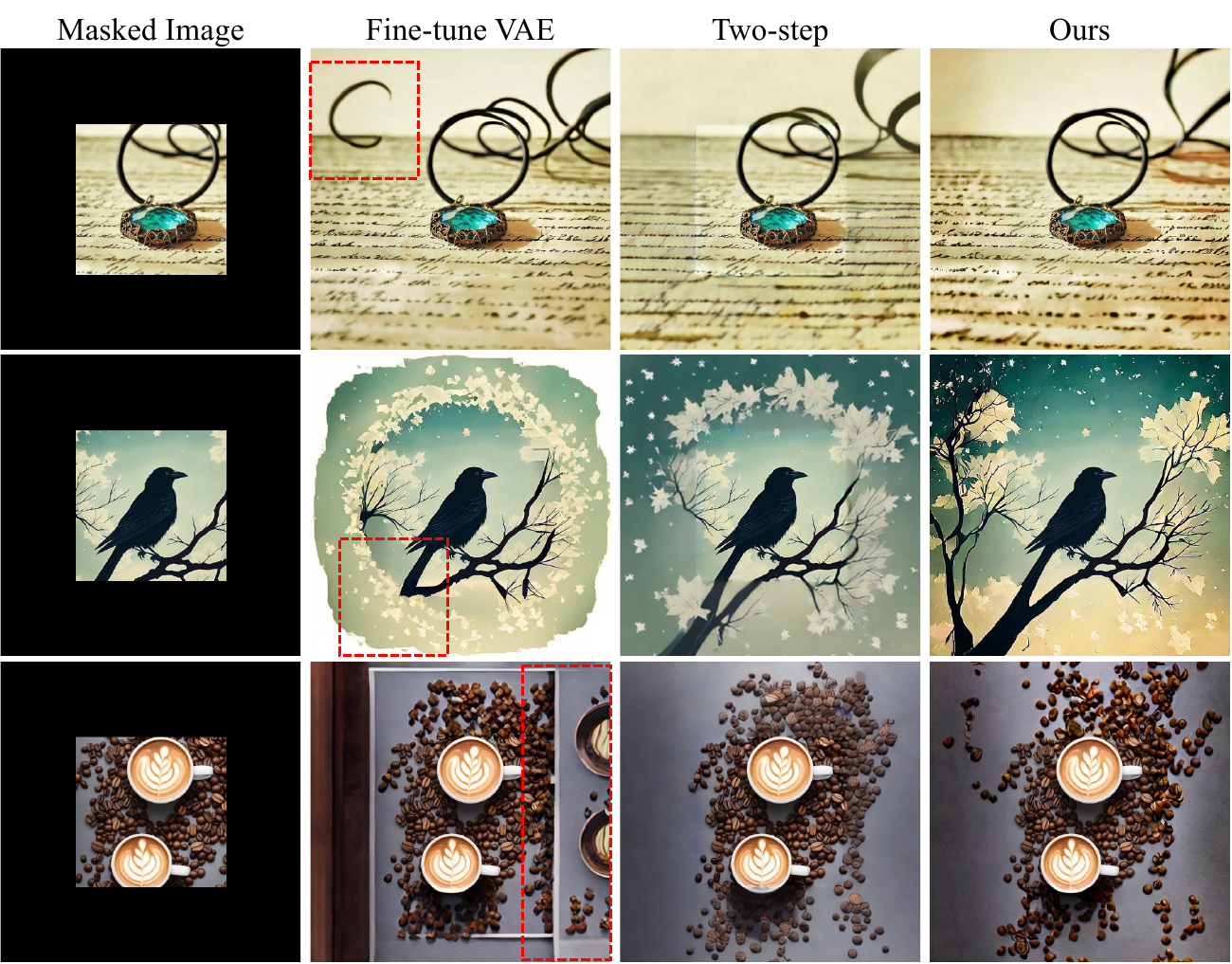}
	\caption{Ablation study for outpainting.}
	\label{fig:ablation_outpainting}
\end{figure}

As shown in ~\cref{fig:ablation_inpainting}, if only the two-step training paradigm is used, noticeable color discontinuities arise, especially in cases where the mask covers a solid-color background (as seen in rows 1 and 2). On the other hand, if only the VAE refinement is applied, the generated content achieves color continuity but often appears sparse or lacks coherent integration with the original image in terms of structure and detail. Our proposed method, which combines both approaches, produces images with strong consistency in both color and content. This combined approach ensures seamless blending, where the inpainted regions not only match the color of the original image but also align with its structural and contextual requirements. \cref{tab:ablation} shows the quantitative results of the components in our method. It can be observed that each component positively contributes to the final results.

\begin{table}[t]
	\centering
	\begin{tabular}{l c c c}
		\hline
		Metrics & \multicolumn{2}{c}{Image Quality} & Consistency \\
		\hline
		Models & FID($\downarrow$) & AS($\uparrow$) & CD($\downarrow$) \\
		\hline
		Fine-tune VAE & 16.835 & 5.725 & 15.354 \\
		Two-step training & 17.081 & 5.576 & 16.080 \\
		Ours  & \textbf{15.434} & \textbf{5.991} & \textbf{8.538} \\
		\hline
	\end{tabular}
	\caption{Ablation study of our methods based on BrushBench.}
	\label{tab:ablation}
\end{table}

We also conduct ablation studies on the outpainting task, as shown in ~\cref{fig:ablation_outpainting}. The results are nearly same as those observed in inpainting: VAE fine-tuning primarily improves color continuity, while the two-step training enhances the coherence between the generated content and the original image.

\section{Conclusion}

In this work, we address the discontinuity issue in diffusion-based image inpainting/outpainting, a problem that has hindered seamless blending between generated and original image regions. We identified and analyzed the underlying causes of the discrepancy and proposed two key solutions. VAE fine-tuning ensures color continuity by refining the model’s reconstruction of masked regions, while the two-step training paradigm simulates the blending operation during training, resulting in smoother blending for image content. Extensive experiments demonstrate that the effectiveness of our approaches. Our methods provide a robust foundation for high-quality image inpainting/outpainting and have the potential to be widely applied in tasks requiring seamless visual integration.

\newpage
{
	\small
	\bibliographystyle{ieeenat_fullname}
	\bibliography{main}
}

\end{document}